%% file: main.tex
\documentclass[10pt,twocolumn,letterpaper]{article}
\usepackage{iccv}
\usepackage{epsfig}
\usepackage{graphicx}
\usepackage{amsmath}
\usepackage{amssymb}

\usepackage{subcaption}
\usepackage{multirow}
\usepackage{multicol}
\usepackage{floatrow}
\usepackage{algorithm}
\usepackage{algorithmic}
\usepackage{booktabs}
\usepackage{mathrsfs, eucal}
\newcommand{\conditioner}{\tau_\theta}
\newcommand{\R}{\mathbb{R}}
\usepackage[pagebackref=true,breaklinks=true,letterpaper=true,colorlinks,bookmarks=false]{hyperref}

\input CSMacrosV2.tex

\usepackage[pagebackref=true,breaklinks=true,letterpaper=true,colorlinks,bookmarks=false]{hyperref}

\newcommand{\diffmask}{\textup{DiffuMask}\xspace}
\newcommand{\Ours}{\textup{DiffuMask}\xspace}

\usepackage{array}
\newcolumntype{I}{!{\vrule width 3pt}}
\newlength\savedwidth

\newlength\savewidth
\newcommand\shline{\noalign{\global\savewidth\arrayrulewidth
                           \global\arrayrulewidth 0.5pt}%
                  \hline
                  \noalign{\global\arrayrulewidth\savewidth}}

\DeclareMathOperator*{\argmax}{arg\,max}

\usepackage[breaklinks=true,bookmarks=false]{hyperref}
\hypersetup{
colorlinks=true,
linkcolor=black
}

\usepackage{tabulary,multirow,overpic,xcolor,subfloat}

\definecolor{linkcolor}{HTML}{ED1C24}

\newcommand{\app}{\raise.17ex\hbox{$\scriptstyle\sim$}}

\def\x{\times}
\newcolumntype{x}[1]{>{\centering\arraybackslash}p{#1pt}}
\newcolumntype{y}[1]{>{\raggedright\arraybackslash}p{#1pt}}

\definecolor{Gray}{gray}{0.5}
\newcommand{\tablestyle}[2]{\setlength{\tabcolsep}{#1}\renewcommand{\arraystretch}{#2}\centering\footnotesize}
\makeatletter\renewcommand\paragraph{\@startsection{paragraph}{4}{\z@}
  {.5em \@plus1ex \@minus.2ex}{-.5em}{\normalfont\normalsize\bfseries}}\makeatother

\newcommand{\lmatch}[1]{{\cal L}_{\rm match}(#1)}

\def\eg{\emph{e.g.}}

\def\ie{\emph{i.e.}}

\newcommand \footnoteONLYtext[1]
{
	\let \mybackup \thefootnote
	\let \thefootnote \relax
	\footnotetext{#1}
	\let \thefootnote \mybackup
	\let \mybackup \imareallyundefinedcommand
}

\iccvfinalcopy 

\def\iccvPaperID{4328} 

\ificcvfinal\fi

\begin{document}

\title{\Ours: 
Synthesizing Images with Pixel-level Annotations for Semantic Segmentation Using Diffusion Models}

\author{Weijia Wu$^{1,3}$,
~~
Yuzhong Zhao$^2$,
~~
Mike Zheng Shou$^3$\footnote{$^*$ Corresponding author},
~~
Hong Zhou$^1$$^*$,
~~
Chunhua Shen$^{1,4}$\\[0.205cm]
\normalsize 
$^1$ Zhejiang University~~
$^2$ University of Chinese Academy of Sciences~~
$^3$ National University of Singapore ~~
$^4$ Ant Group
}

\input{figures/fig1}

 \maketitle
 \ificcvfinal\thispagestyle{empty}\fi

\begin{abstract}
\footnoteONLYtext{$^*$ Corresponding author}
Collecting and annotating images with pixel-wise labels is time-consuming and laborious.
%
In contrast, synthetic data can be freely available using a generative model~(\eg, DALL-E, Stable Diffusion).
%
%
%
In this paper, 
we show that it is possible to automatically obtain accurate semantic masks of synthetic images generated by the 
Off-the-shelf 
Stable Diffusion model, which uses only text-image pairs during training. 
%
Our approach, 
termed 
\Ours, exploits the potential of the cross-attention map between text and image, which is natural and seamless to extend the text-driven image synthesis to semantic mask generation.
\Ours uses text-guided cross-attention information to localize class/word-specific regions, which are combined with practical techniques to create a novel high-resolution and class-discriminative pixel-wise mask.
The methods help to 
significantly 
reduce data collection and annotation costs.
%
%
Experiments demonstrate that the existing segmentation methods trained on synthetic data of \Ours can achieve a competitive performance over the counterpart of real data (VOC 2012, Cityscapes).
For some classes (\eg{}, bird), \Ours presents 
promising performance, close to the 
state-of-the-art 
result of real data (
within \textbf{3\%} mIoU gap).
Moreover, in the open-vocabulary segmentation (zero-shot) setting, \Ours achieves new 
state-of-the-art 
results on the \texttt{Unseen} classes of VOC 2012.
The project website can be found at \href{https://weijiawu.github.io/DiffusionMask/}{\color{blue}{$\tt DiffuMask$}}.
\end{abstract}

\section{Introduction}
Semantic segmentation is a fundamental task in vision, and existing data-hungry semantic segmentation models usually require a large amount of data with pixel-level annotations to achieve significant progress. 
Unfortunately, pixel-wise mask annotation is a labor-intensive and expensive process.
For example, labeling a single semantic urban image in Cityscapes~\cite{cordts2016cityscapes} can take up to 60 minutes, underscoring the level of difficulty involved in this task
%
%
Additionally, in some cases, it may be challenging or even impossible to collect images due to existing privacy and copyright.
To reduce the cost of annotation, weakly-supervised learning has become a popular approach in recent years. This approach involves training strong segmentation models using weak or cheap labels, such as image-level labels~\cite{ahn2018learning,lee2021anti,wu2021embedded,xu2021leveraging,ru2021learning,ru2022learning}, points~\cite{akiva2021towards}, scribbles~\cite{lin2016scribblesup,zhang2021affinity}, and bounding boxes~\cite{lee2021bbam}.
Although these methods are free of pixel-level annotations, still suffer from several disadvantages, including low-performance accuracy, complex training strategy, indispensable extra annotation cost (\eg{}, edge), and image collection cost.

With the great development of computer graphics~(\eg{}, generative model), an alternative way is to utilize synthetic data, which is largely available from the virtual world, and the pixel-level ground truth can be freely and automatically generated.
DatasetGAN~\cite{zhang2021datasetgan} firstly exploits the feature space of a trained GAN and trains a shallow decoder to produce pixel-level labeling.
BigDatasetGAN~\cite{li2022bigdatasetgan} extends DatasetGAN to handle the large class diversity of ImageNet.
However, both methods suffer from certain drawbacks, the need for a small number of \textbf{pixel-level} labeled examples to generalize to the rest of the latent space and suboptimal performance due to imprecise generative masks.

Recently, large-scale language-image generation (LLIG) models, such as DALL-E~\cite{ramesh2022hierarchical}, and Stable Diffusion~\cite{rombach2022high}, have shown phenomenal generative semantic and compositional power, as shown in Fig.~\ref{fig:teaser}. 
Given one language description, the text-conditioned image generation model can create corresponding semantic things and stuff,
where visual and textual embedding are fused using spatial cross-attention.  
We dive deep into the cross-attention layers and explore how they affect the generative semantic object and structure of the image.
We find that cross-attention maps are the core, which binds visual pixels and text tokens of the prompt text.
%
Also, the cross-attention maps contain rich class~(text token) discriminative spatial localization information, which critically affects the generated image.

\begin{figure}
	\begin{minipage}{0.98\linewidth}
		\includegraphics[width=0.99\linewidth]{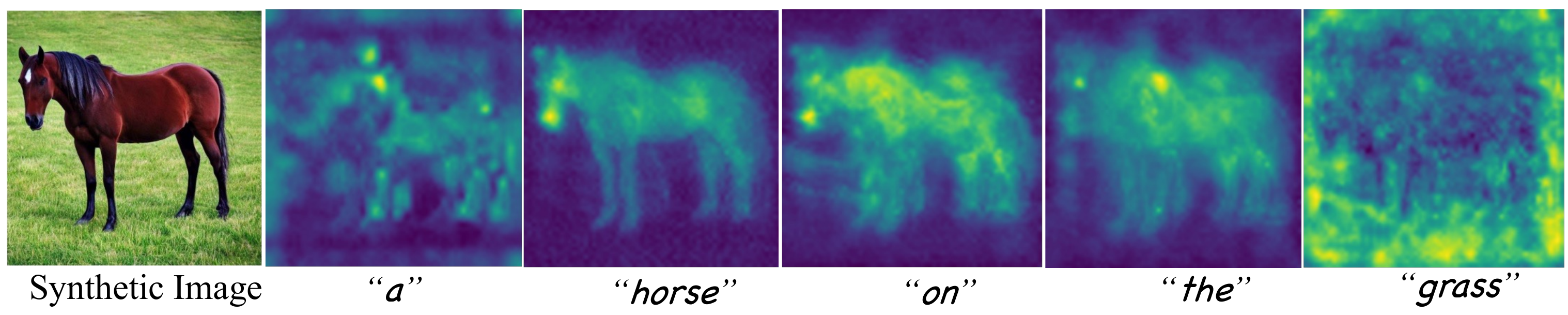}
        \vspace{-0.2cm}
		\subcaption{Cross attention maps of different text tokens.}
		\label{fig:1a}	
	\end{minipage}
	\qquad
	\begin{minipage}{0.98\linewidth}
		\centering
		\includegraphics[width=0.99\linewidth]{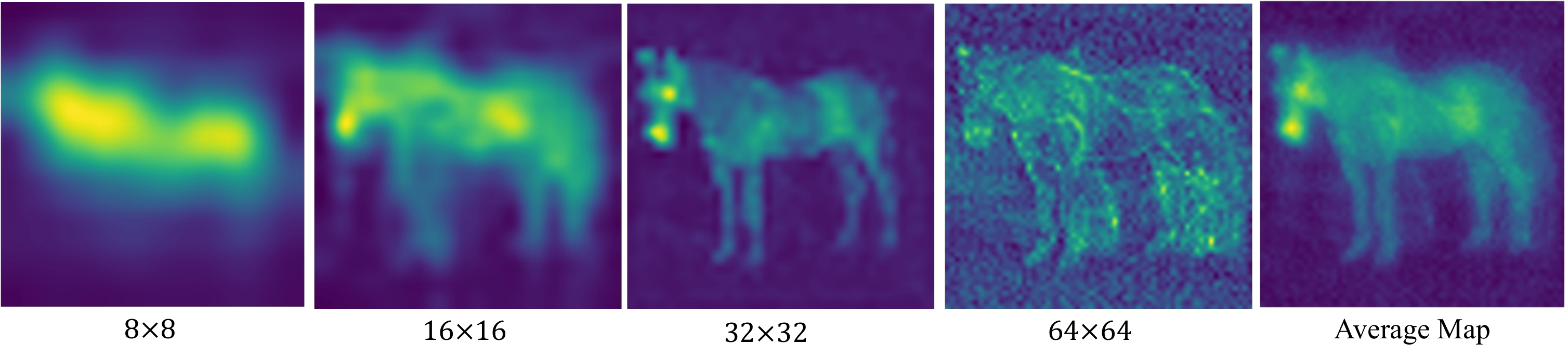}
        \vspace{-0.2cm}
		\subcaption{Cross attention maps of different resolutions.}
		 \label{fig:1b}	
	\end{minipage}
        \qquad
	\begin{minipage}{0.98\linewidth}
		\centering
		\includegraphics[width=0.99\linewidth]{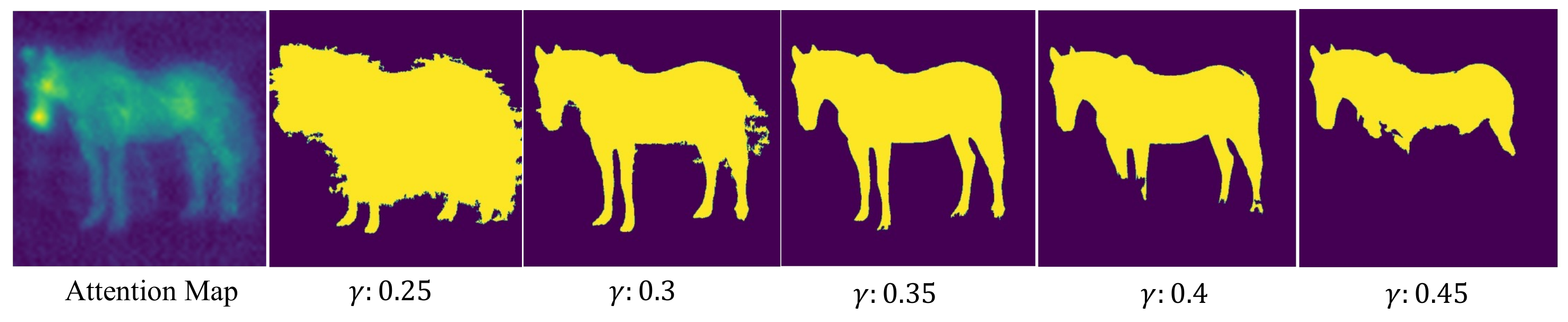}
        \vspace{-0.2cm}
	\subcaption{Binarization 
 Mask with different thresholds $\gamma$ in Equ.~\eqref{eq:binarization}.}
		 \label{fig:1c}	
	\end{minipage}
        \vspace{-0.2cm}
	\caption{\textbf{Cross-attention maps of a text-conditioned diffusion model~(\ie{}, Stable Diffusion~\cite{rombach2022high}).} Prompt language: `\texttt{a horse on the grass}'.
}
\label{fig2_attention}
\end{figure}

\textbf{Can the attention map be used as mask annotation?}
Consider semantic segmentation~\cite{(voc)everingham2010pascal,cordts2016cityscapes}---a ‘good’ pixel-level semantic mask annotation should satisfy two conditions: (a) class-discriminative (\ie{}, localize and distinguish the categories in the image); (b) high-resolution, precise mask (\ie{}, capture fine-grained detail).
Fig.~\ref{fig:1b} presents a visualization of cross attention map between text token and vision.
$8\times8$, $16\times16$, $32\times32$, and $64\times64$, as four different resolutions, are extracted from different layers of the U-Net of Stable Diffusion \cite{rombach2022high}.
$8\times8$ feature map is the lowest resolution, including obvious class-discriminative location.
$32\times32$ and $64\times64$ feature maps include high-resolution and highlight fine-grained details.
The average map shows the possibility for us to use for semantic segmentation, where it is class-discriminative and fine-grained.
To further validate the potential of the attention map of the generative task, we convert the probability map to a binary map with fixed thresholds $\gamma$, and refine them with Dense CRF~\cite{krahenbuhl2011efficient}, as shown in Fig.~\ref{fig:1c}.
With the $0.35$ threshold, the mask presents a wonderful precision on fine-grained details~(\eg{}, foot, ear of the `\textit{horse}').

%
%
Based on the above observation, we present \Ours, an automatic procedure to generate a massive high-quality image with a pixel-level semantic mask.
%
Unlike DatasetGAN~\cite{zhang2021datasetgan} and BigDatasetGAN~\cite{li2022bigdatasetgan}, \diffmask does not require any pixel-level annotations. This approach takes full advantage of powerful zero-shot text-to-image generative models such as Stable Diffusion~\cite{rombach2022high}, which are trained on web-scale image-text pairs.
%
%
\Ours mainly includes two advantages for two challenges: 1) \textit{Precise Mask.} An adaptive threshold of binarization is proposed to convert the probability map~(attention map) to a binary map, as the mask annotation.
Besides, noise learning~\cite{northcutt2021confident,song2022learning} is used to filter noisy labels.
2) \textit{Domain Gap:} retrieval-based prompt~(various and verisimilar prompt guidance) and data augmentations~(\eg{}, Splicing~\cite{bochkovskiy2020yolov4}), as two effective solutions, are designed to reduce the domain gap via enhancing the diversity of data. 
With the above advantages, \diffmask can generate infinite images with pixel-level annotation for any class without human effort.
These synthetic data can then be used for training any semantic segmentation architecture~(\eg{}, mask2former~\cite{cheng2022masked}), replacing real data.  
To 
summarize, our contributions are three-folds:
\begin{itemize}
    \itemsep-0.1cm 
    \item We show a novel insight that it is possible to automatically obtain the synthetic image and mask annotation from a text-supervised pre-trained diffusion model. 
    
    %
    
    \item We present \diffmask, an automatic procedure to generate massive image and pixel-level semantic annotation \textit{without} human effort and any manual mask annotation,
    which exploits the potential of the cross-attention map between text and image.
    
    \item Experiments demonstrate that  segmentation methods trained on \Ours perform competitively on real data, \eg{}, VOC 2012.
    For some classes, \eg{}, \texttt{dog}, the performance is close to that of training with real data 
    (within \textbf{3\%} gap).
    Moreover, in the open-vocabulary segmentation (zero-shot) setting, \Ours achieves  new SOTA results on the \texttt{Unseen}  classes of VOC 2012.

\end{itemize}

\section{Related Work}

\textbf{Reducing Annotation Cost.} Various ways can be explored to reduce the segmentation data cost, including interactive human-in-the-loop annotation~\cite{acuna2018efficient,ling2019fast}, nearest-neighbor mask transfer~\cite{guillaumin2014imagenet}, or weak/cheap mask annotation supervision in different levels, such as image-level labels~\cite{ahn2018learning,lee2021anti,wu2021embedded,xu2021leveraging,ru2021learning,ru2022learning}, points~\cite{akiva2021towards}, scribbles~\cite{lin2016scribblesup,zhang2021affinity},  and bounding boxes~\cite{lee2021bbam,chen2014beat,kulharia2020box2seg}.
Among the above-related works, image-level label supervised learning~\cite{ru2021learning,ru2022learning} presents the lowest cost, and its performance is unacceptable.
Bounding boxes~\cite{chen2014beat,kulharia2020box2seg} annotation usually shows a competitive performance than pixel-wise supervised methods, but its annotation cost is the most expensive.
By comparison, synthetic data presents many advantages, including lower data cost without image collection, and infinite availability for enhancing the diversity of data.

\textbf{Image Generation.}
Image generation is a basic and challenging  task in computer vision.
There are several mainstream methods for the task, including Generative Adversarial Networks
(GAN)~\cite{goodfellow2020generative}, Variational autoencoders (VAE)~\cite{kingma2013auto}, flow-based models~\cite{dinh2014nice}, and Diffusion Probabilistic Models (DM)~\cite{sohl2015deep,rombach2022high,gu2023mix}.
Recently, the diffusion model has drawn lots of attention due to its wonderful performance.
GLIDE~\cite{nichol2021glide} used pre-trained  language model~(CLIP~\cite{radford2021learning}) and the  cascaded diffusion structure for text-to-image  generation.
Similarly, DALL-E 2~\cite{ramesh2022hierarchical} of OpenAI Imagen~\cite{saharia2022photorealistic} obtain the corresponding text embedding with CLIP and adopted a similar hieratical  structure to generate images.
To increase accessibility and reduce  significant resource consumption, Stable Diffusion~\cite{rombach2022high} of Stability AI introduced a novel direction in which the model diffuses on VAE latent spaces instead of pixel spaces. 

\textbf{Synthetic Dataset Generation.}
Prior works~\cite{kar2019meta,devaranjan2021unsupervised} for dataset synthesis mainly utilize 3D scene graphs to render images and their labels.
2D methods, \ie{}, Generative Adversarial Networks (GAN)~\cite{goodfellow2020generative} mainly is used to solve domain adaptation task~\cite{choi2019self,choi2019self}, which leverages 
image-to-image translation to reduce the domain gap.
Recently, inspired by the success of generative model~(\eg{}, DALL-E 2, Stable Diffusion), some works further try to explore the potential of synthetic data to replace real data as the training data in many downstream tasks, including image classification~\cite{he2022synthetic,besnier2020dataset}, object detection~\cite{wu2022synthetic,ni2022imaginarynet,ge2022dall,ge2022neural,zhao2023generative,zhao2023flowtext}, image segmentation~\cite{li2022bigdatasetgan,zhang2021datasetgan,li2023guiding}, 
3D Rendering~\cite{zhang2020image,poole2022dreamfusion}.
DatasetGAN~\cite{zhang2021datasetgan} utilized a few labeled real images to train a segmentation mask decoder, leading to an infinite synthetic image and mask generator.
Based on DatasetGAN, BigDatasetGAN~\cite{li2022bigdatasetgan} scale the class diversity to ImageNet size, which generates 1k classes with manually annotated 5 images per class.
With Stable diffusion and Mask R-CNN pre-trained on COCO dataset, Li \textit{et al.}~\cite{li2023guiding} design and train a grounding module to generate images and segmentation masks.
Different from the above methods, we go one step further and synthesize accurate semantic labels by exploiting the potential of cross attention map between text and image. 
%
%
One significant advantage of the \diffmask is that it does not require any manual localization annotations~(\ie{}, box and mask) and only rely on \textit{text supervision}.


\section{Methodology}
In this paper, we explore simultaneously generating images and the semantic mask described in the text prompt with the existing pre-trained diffusion model.
Using the synthetic data to train the existing segmentation methods, and apply them to the real images.

The core is to exploit the potential of the \textit{cross-attention map} in the generative model and \textit{domain gap} between synthetic and real data, providing corresponding new insights, solutions, and analysis.
We introduce the preliminary of cross attention in Sec.~\ref{Problem}, 
Mask generation and refinement with cross-attention map in text-conditioned diffusion models in Sec.~\ref{Mask}, data diversity enhancement with prompt engineering in Sec.~\ref{Prompt}, data augmentation in Sec.~\ref{Data}.

\subsection{Cross-Attention of Text-Image}
\label{Problem}
Text-guided generative models~(\eg{}, Imagen~\cite{saharia2022photorealistic}, Stable Diffusion~\cite{rombach2022high}) use a text prompt $\mathcal{P}$ to guide the content-related image $\mathcal{I}$ generation from a random gaussian image noise $z$, where visual and textual embedding are fused using the spatial cross-attention.  
Specifically, Stable Diffusion~\cite{rombach2022high} consists of a text encoder, a variational autoencoder~(VAE), and a U-shaped network~\cite{ronneberger2015u}.
The interaction between the text and vision occurs in the U-Net for the latent vectors at each time step, where cross-attention layers are used to fuse the embeddings of the visual and textual features and produce spatial attention maps for each textual token.
Formally, for step $t$, the visual features of the noisy image $\varphi(z_t) \in \R^{H \times W \times C}$ are flatted and linearly projected into a \texttt{Query} vector $Q = \ell_Q(\varphi(z_t))$.
The text prompt $\mathcal{P}$ is projected into the textual embedding $\conditioner(\mathcal{P}) \in \R^{N\times d }$ ($N$ refers to the sequence length of text tokens and $d$ is the latent projection dimension) with the text encoder~$\conditioner$,
then is mapped into a \texttt{Key} matrix $K = \ell_K(\conditioner(\mathcal{P}))$ and a \texttt{Value} matrix $V = \ell_V(\conditioner(\mathcal{P}))$, via learned projections $\ell_Q, \ell_K, \ell_V$.
The \textit{cross attention maps} can be calculated by: 
\begin{equation}
\mathcal{A}=\text{Softmax}\left(\frac{QK^T}{\sqrt{d}}\right),
\label{equ1}
\end{equation}
where $\mathcal{A} \in \R^{H\times W \times N}$ (re-shape). 
For $j$-th text token, \eg{}, \textit{horse} on Fig.~\ref{fig:1a}, the corresponding weight $\mathcal{A} _{j} \in \R^{H\times W}$ on the visual map $\varphi(z_t)$ can be obtained. 
Finally, the output of cross-attention can be obtained with $\widehat{\varphi}\left(z_t\right)=\mathcal{A} V$, which is then used to update the spatial features $\varphi(z_t)$.
%



\subsection{Mask Generation and Refinement}
Based on Equ.~\ref{equ1}, we can obtain the corresponding cross attention map $\mathcal{A} _{j}^{s,t}$.
$s$ denotes the attention map from $s$-th layer of U-Net, and corresponding to four different resolutions, \ie{}, $8\times8$, $16\times16$, $32\times32$, and $64\times64$, as shown in Fig.~\ref{fig:1b}.
$t$ denotes $t$-th diffusion step~(time).
Then the average cross-attention map can be calculated by aggregating the multi-layer and multi-time attention maps as follows:
\begin{align}
    \mathcal{\hat{A}}_j = \frac{1}{S\cdot T}\sum_{s\in S,t\in T}\frac{\mathcal{A} _{j}^{s,t}}{\text{max}(\mathcal{A} _{j}^{s,t})},\label{eq:agg}
\end{align}
where $S$ and $T$ refer to the total steps and the number of layers (\ie{}, four for U-Net). 
Normalization is necessary due the value of the attention map from the output of \texttt{Softmax} is not a probability between 0 and 1.

\label{Mask}
\subsubsection{Standard Binarization}
\begin{figure}[!t]
	\includegraphics[width=0.99\linewidth]{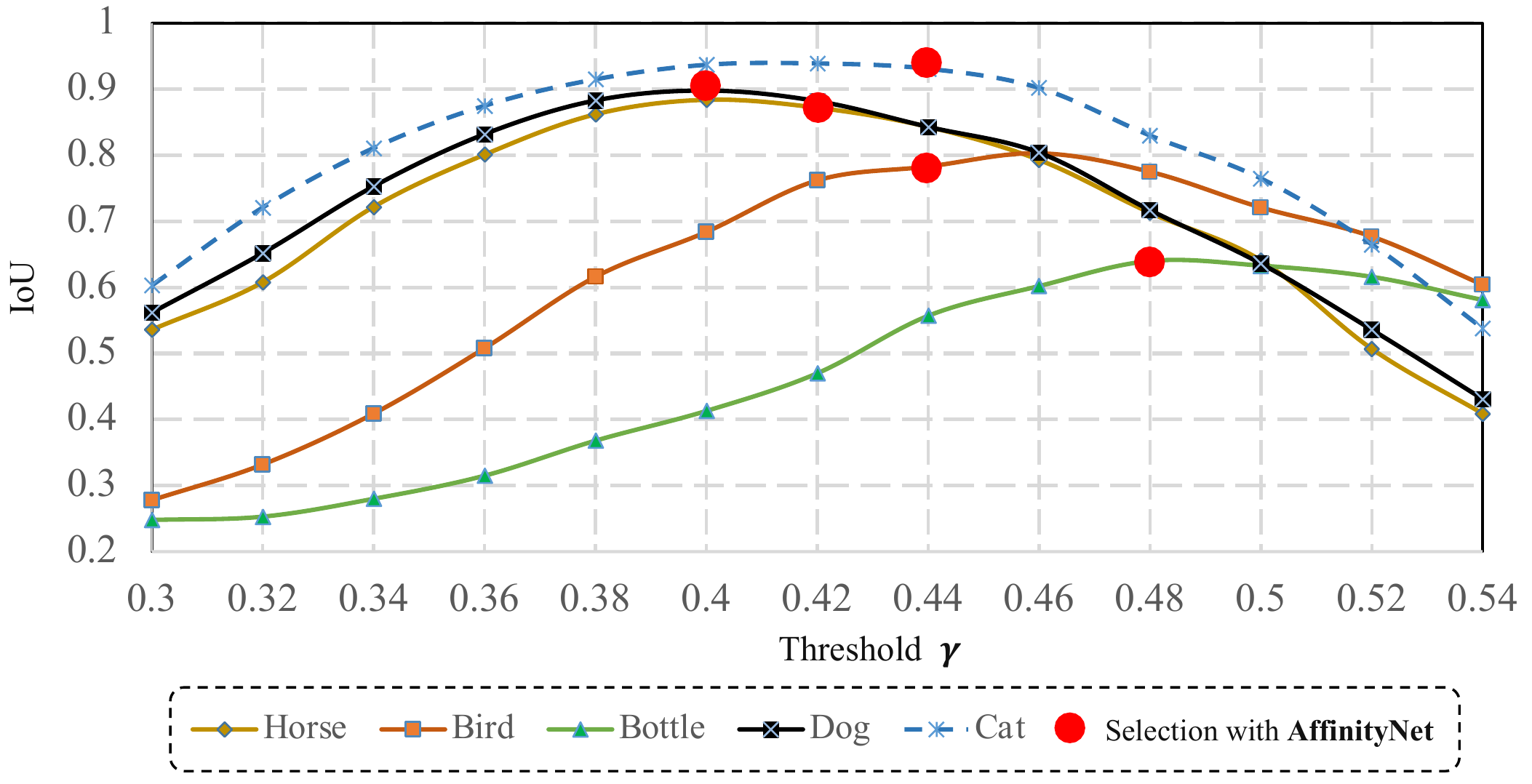}
	\caption{\textbf{Relationship between mask quality~(IoU) and threshold for 
 various 
 categories.} 
 $1k$ generative images are used for each class from Stable Diffusion~\cite{rombach2022high}. Mask2former~\cite{cheng2022masked} pre-trained on Pascal-VOC 2012~\cite{(voc)everingham2010pascal} is used to generate the ground truth. The optimal threshold of different classes usually is different.}
 \vspace{-0.2cm}
\label{iou}
\end{figure}

Given an average attention map~(a  probability map) $M \in \R^{H \times W}$ for $j$-th text token produced by the cross attention in Equ.~\eqref{equ1},
it is essential to convert it to a binary map, where pixels with $1$ as the foreground region~(\eg{}, `\texttt{horse}').
Usually, as shown in Fig.~\ref{fig:1c}, the simplest solution for the binarization process is using a fixed threshold value $\gamma$, and refining with DenseCRF~\cite{krahenbuhl2011efficient} (local relationship 
defined by 
color 
and distance of pixels) as follows:
\begin{equation}
    B=
    \text{DenseCRF}(\left [ \gamma;\mathcal{\hat{A}}_j  \right ]_{\texttt{argmax}} )\;.
    \label{eq:binarization}
\end{equation}
The above method is not practical and effective, while the \textit{optimal threshold} of each image and each category are not exactly the same.
To explore the relationship between threshold and binary mask quality, we set a simple analysis experiment.
Stable Diffusion~\cite{rombach2022high} is used to generate 1k images and corresponding attention maps for each class.
The prediction of Mask2former~\cite{cheng2022masked} pre-trained on Pascal-VOC 2012 as the ground truth is adopted to calculate the quality of mask quality~(mIoU), as shown in Fig.~\ref{iou}. 
The optimal threshold of different classes usually are different, \eg{}, around $0.48$ for `\texttt{Bottle}' class, different from that~(\ie{},  around $0.39$) of `\texttt{Dog}' class.
To achieve the best quality of the mask, the \textit{adaptive threshold} is a feasible solution for the various binarization for each image and class.

\begin{figure*}[t]
	\includegraphics[width=0.99\linewidth]{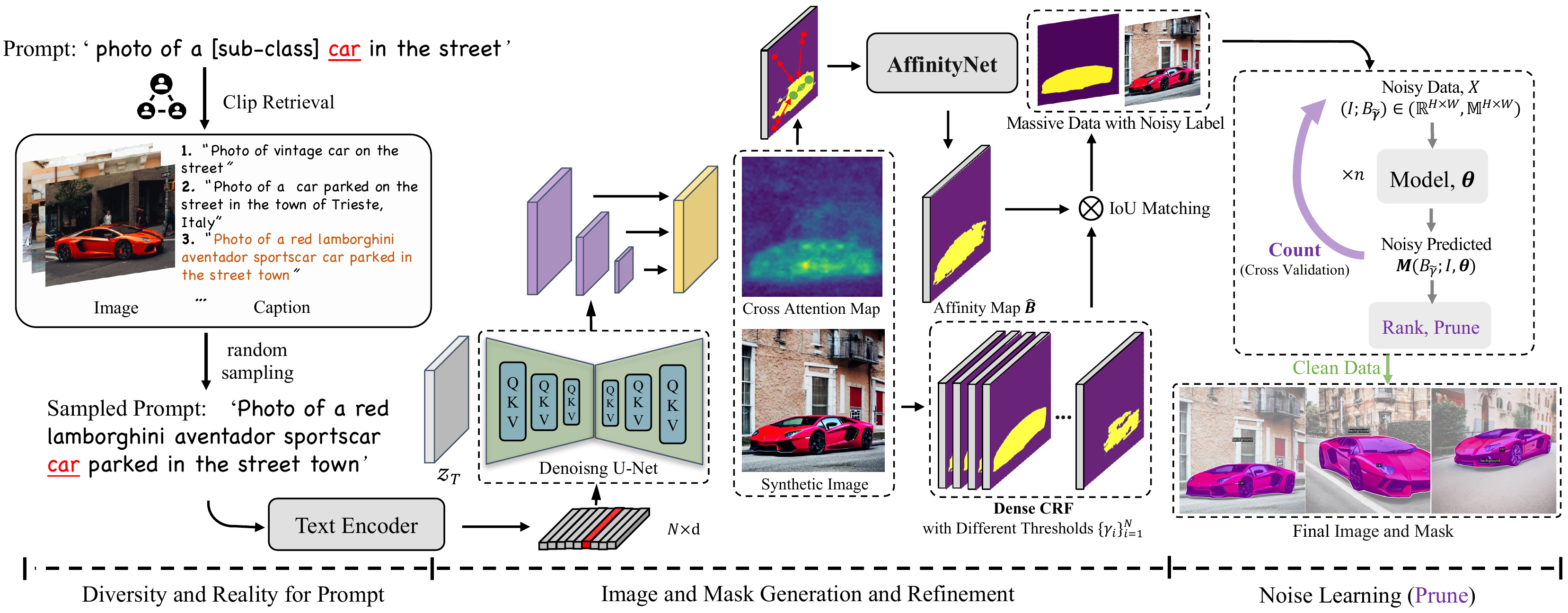}
	\vspace{-0.05cm}
	\caption{\textbf{Pipeline for \diffmask with a prompt: `\texttt{Photo of a [sub-class] car in the street}'}. \diffmask mainly includes three steps: 1) Prompt engineering is used to enhance the diversity and reality of prompt language (Sec.~\ref{Prompt}). 2) Image and mask generation and refinement with adaptive threshold from AffinityNet~(Sec.~\ref{Mask}). 3) Noise learning is designed to further improve the quality of data via filtering the noisy label~(Sec.~\ref{CL}). 
    }
    \vspace{-0.25cm}
\label{pipeline}
\end{figure*}
\vspace{-0.25cm}
\subsubsection{Adaptive Threshold for Binarization}
It is challenging to determine the optimal threshold for binarizing the probability maps because of the variation in shape and region for each object class.
The image generation relies on \textbf{text-supervision}, which does not provide a precise definition of the shape and region of object classes. 
%
%
For example, the masks with $0.45 
\gamma$ and that with $0.35 
\gamma$ in Fig.~\ref{fig:1c}, the model can not judge which one is better, while no location information as supervision and reference is provided by human effort.

Looking deeper at the challenge, pixels with a middle confidence score cause uncertainty, while that with a high and low score usually represent the true foreground and the background.
To address the challenge, semantic affinity learning (\ie{},  AffinityNet~\cite{ahn2018learning}) is used to give an estimation for those pixels with a middle confidence score.
Thus we can obtain the definition for global prototype, \ie{}, \textit{which semantic masks with different threshold $\gamma$ is suitable to represent the whole prototype}. 
AffinityNet aims to predict semantic affinity between a pair of adjacent coordinates.
During the training phase, those pixels in the middle score range are considered as \textit{neutral}.
If one of the adjacent coordinates is \textit{neutral}, the network simply ignores the pair during training.
Without \textit{neutral} pixels, the affinity label of two coordinates is set to $1$~(positive pair) if their classes are the same, and 0~(negative pair)  otherwise.
During the inference phase, a coarse affinity map $\hat{B} \in \R^{H \times W}$ can be predicted by AffinityNet for each class of each image.
$\hat{B}$ is used to search for a suitable threshold $\hat{\gamma}$ during a search space $\Omega = \{\gamma_i\}_{i=1}^{L}$ as follows:
\begin{equation}
\label{eq:matching}
    \hat{\gamma} = \argmax_{\gamma\in\Omega } \sum_{}^{} \lmatch{\hat{B}, B_{\gamma}},
\end{equation}
where $\lmatch{\hat{B}, B_{\gamma}}$ is a pair-wise \emph{matching cost} of IoU between affinity map $\hat{B}$ and a binary map from attention map with threshold $\gamma$.
As a result, an adaptive threshold $\hat{\gamma}$ can be obtained for each image of each class.
The red points in Fig.~\ref{iou} represent the corresponding threshold from matching with the affinity map.
They are usually close to the optimal threshold.

\begin{figure}[t]
	\begin{minipage}{0.47\linewidth}
		\includegraphics[width=0.99\linewidth]{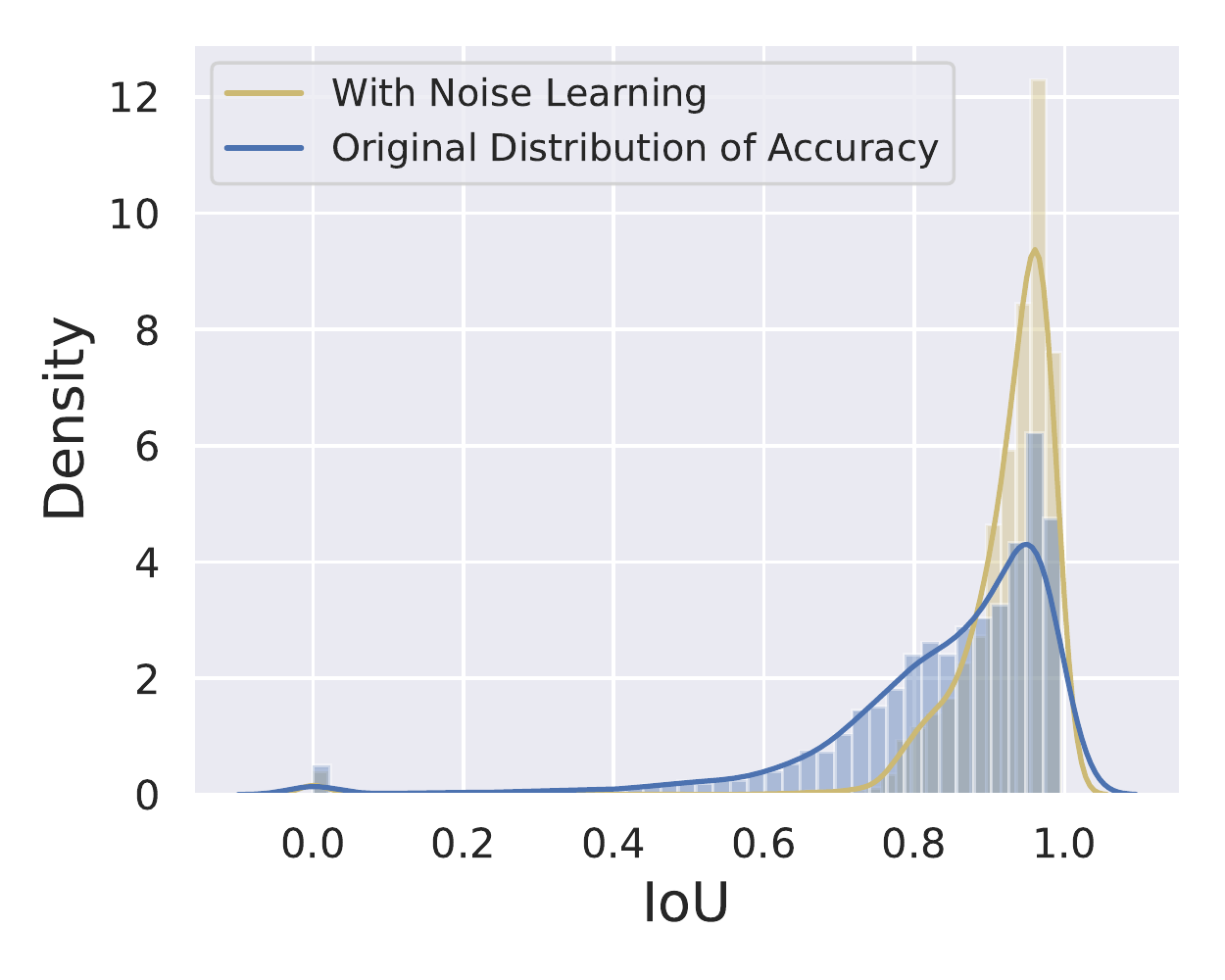}
        \vspace{-0.6cm}
	\subcaption{Distribution of `\texttt{Horse}'.}
		\label{fig:3a}	
	\end{minipage}
    \quad
	\begin{minipage}{0.47\linewidth}
		\centering
		\includegraphics[width=0.99\linewidth]{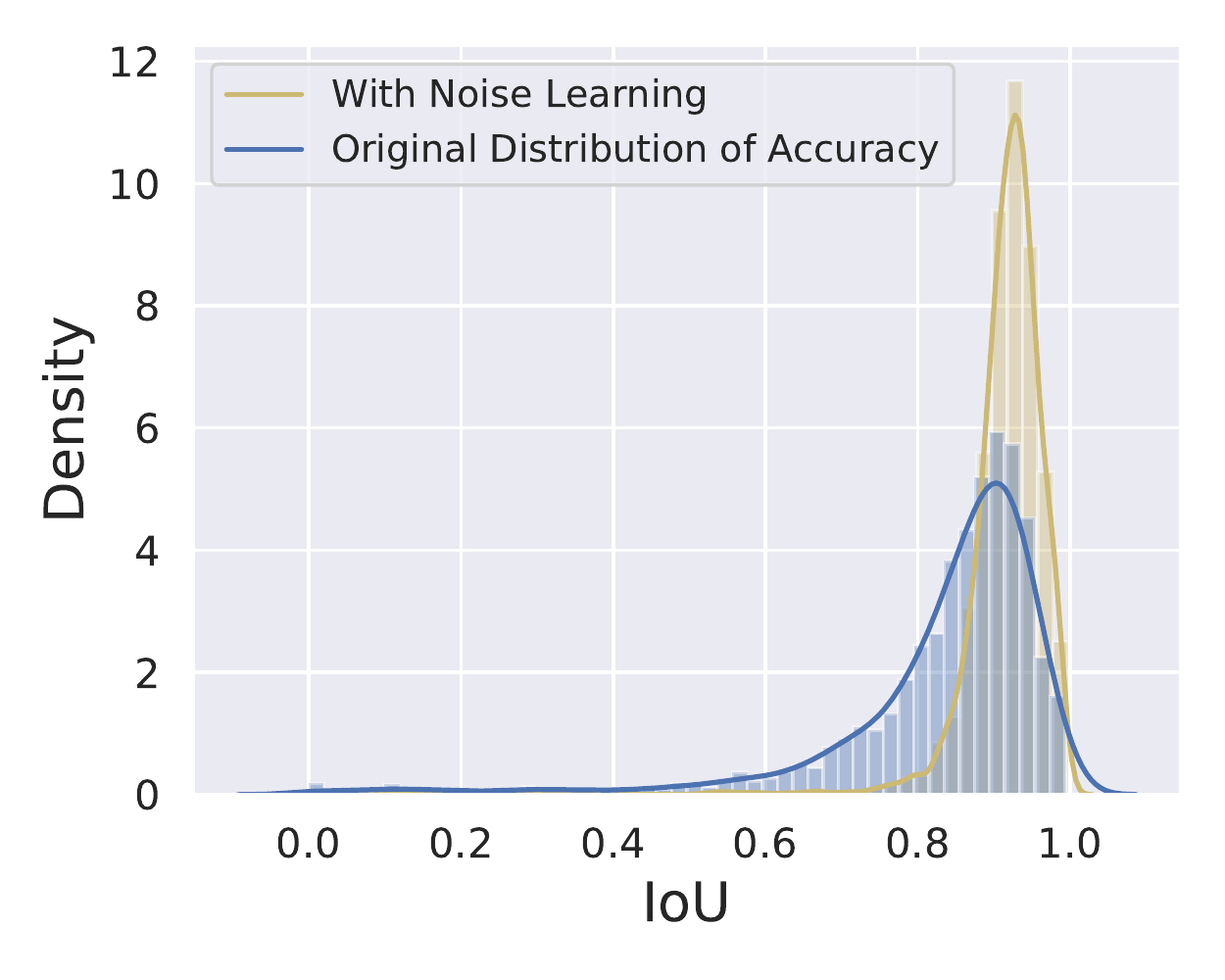}
        \vspace{-0.6cm}
	\subcaption{Distribution of `\texttt{Bird}'.}
		 \label{fig:3b}	
	\end{minipage}
        \vspace{-0.2cm}
	\caption{\textbf{Effect of Noise Learning~(NL).} 30k generative images are used for each class. NL prunes $70\%$ images on the basis of the rank of IoU.
    Mask2former~\cite{cheng2022masked} pre-trained on VOC 2012~\cite{(voc)everingham2010pascal} is used to generate the ground truth. NL brings obvious improvement in mask quality by pruning data.}
\label{iou_density}
\end{figure}

\subsection{Noise Learning}
\label{CL}
Although refined mask $B_{\hat{\gamma}}$ presents a competitive result, there are still existing noisy labels with low precision.
Fig.~\ref{iou_density} provides the probability density distribution of IoU for the `\texttt{Horse}' and `\texttt{Bird}' classes.
The masks with IoU under $80\%$ account for a non-negligible proportion and may cause a significant performance drop.
%
%
Inspired by noise learning~\cite{northcutt2021confident,song2022learning,chen2019understanding} for the classification task, we design a simple, yet effective noise learning~(NL) strategy to prune the noise labels for the segmentation task.

NL improves the data quality by identifying and filtering noisy labels.
The main procedure~(see Fig.~\ref{pipeline}) comprises two steps:  (1) \textbf{Count}: estimating the distribution of label noise ${Q_{B_{\hat{\gamma}}, B^*}}$ to characterize pixel-level label noise, $B^*$ refers to the prediction of model. (2) \textbf{Rank}, and \textbf{Prune}: filter out noisy examples and train with errors removed data.
Formally, given massive generative images and annotations $\{(\mathcal{I},B_{\hat{\gamma}})\}$, a segmentation model $\bm{\theta}$~(\eg{}, Mask2former~\cite{cheng2022masked}, Mask-RCNN~\cite{he2017mask}) is used to predict out-of-sample probabilities of segmentation result $\bm{\theta}: \mathcal{I} \rightarrow \bm{M}_c(B_{\hat{\gamma}}; \mathcal{I}, \bm{\theta})$ by cross-validation.
Then we can estimate the joint distribution of noisy labels $B_{\hat{\gamma}}$ and true labels, ${Q^c_{B_{\hat{\gamma}}, B^*}}=\Phi_{\text{IoU}}(B_{\hat{\gamma}}, B^*)$, where $c$ denotes $c$-th class.
With ${Q^c_{B_{\hat{\gamma}}, B^*}}$, some interpretable and explainable ranking methods, such as loss reweighting~\cite{DBLP:conf/iclr/GoldbergerB17_smodel,NIPS2013_5073} can be used for CL to find label errors using.
In this paper, we adopt a simple and effective modularized rank and prune method, \ie{}, \textit{Prune by Class}, which decouples the model and data cleaning procedure.
For each class, select and prune $\alpha \%$ examples with the lowest self-confidence ${Q^c_{B_{\hat{\gamma}}, B^*}}$ as the noisy data, and train model $\bm{\theta}$ with the remaining clean data.
While $\alpha \%$ is set to $50\%$, the probability density distribution of IoU from the remaining clean data is presented in Fig.~\ref{iou_density} (yellow).
CL can bring an obvious gain for the mask precision, which further taps the potential of attention map as mask annotation.

\begin{figure}[!t]
	\includegraphics[width=0.99\linewidth]{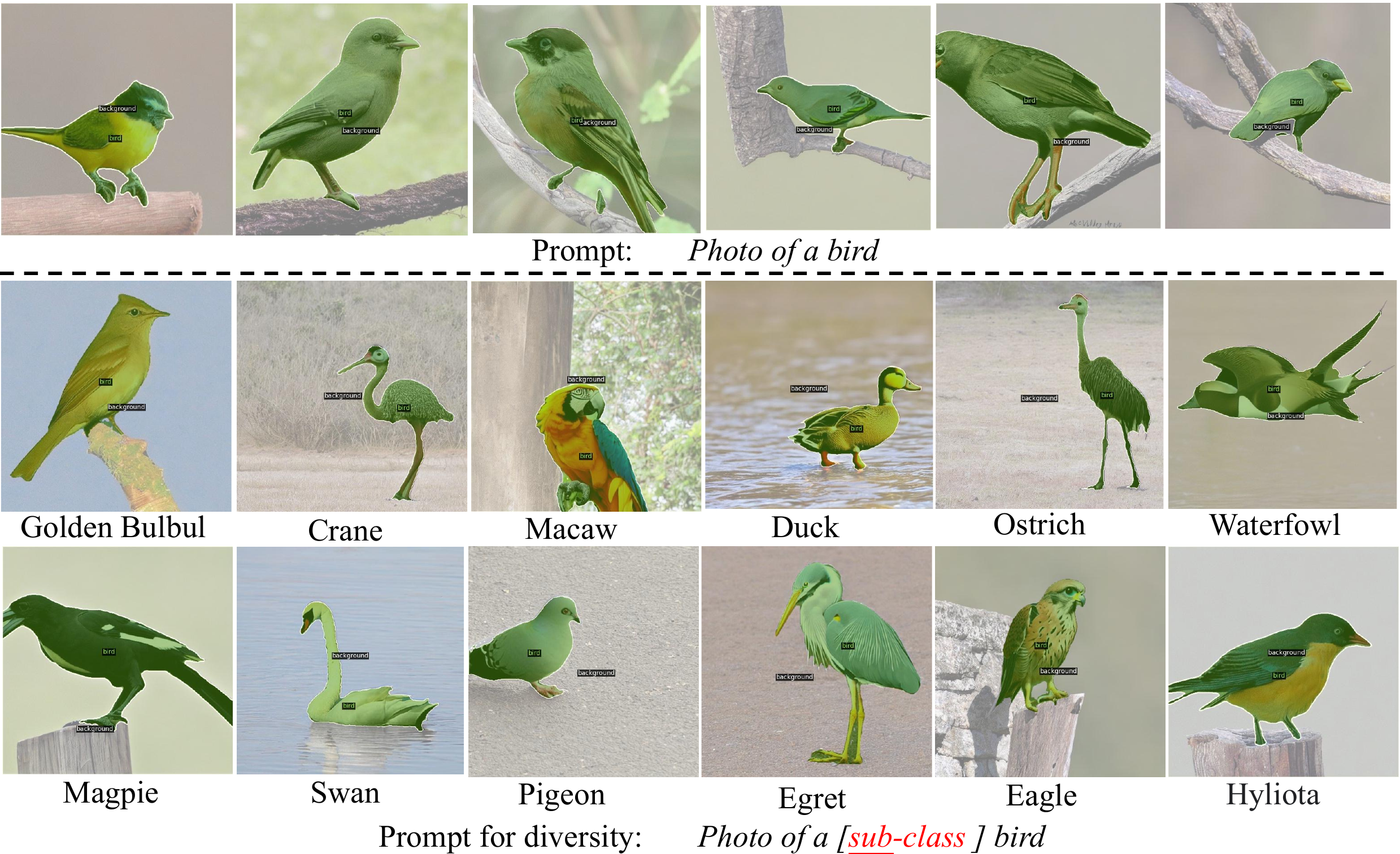}
	\vspace{-0.2cm}
	\caption{\textbf{Prompt for diversity in sub-class for the \textit{bird} class.} $100$ sub-classes for \textit{bird} class in total for our experiment. The same prompt strategy is used for other classes, \eg, cat, car.}
	\vspace{-0.1cm}
\label{VIS12}
\end{figure}

\subsection{Prompt Engineering}
\label{Prompt}
Previous works~\cite{ni2022imaginarynet,witteveen2022investigating} have shown the effectiveness of prompt engineering on diversity enhancement of generative data.
These studies utilize a variety of prompt modifiers to influence the generated images, \eg{}, GPT3 used by ImaginaryNet~\cite{ni2022imaginarynet}.
Unlike generation-based or modification-based prompts, we design two practical, reality-based prompt strategies.

\textbf{Prompt with Sub-Classes.} Simple text prompts, such as `\texttt{Photo of a bird}', often results in monotony for generative images, as depicted in Fig.~\ref{VIS12}~(upper), they fail to capture the diverse range of objects and scenes found in the real world. 
%
To address this challenge, we incorporate `\texttt{sub-classes}' for each category to improve diversity.
To achieve this, we select $K$ sub-classes for each category from Wiki\footnote{https://en.wikipedia.org/wiki/Main\_Page} and integrate this information into the prompt templates.
Fig.~\ref{VIS12}~(down) presents an example for `\texttt{bird}' category.
Given $K$ sub-classes, \ie{}, Golden Bullul, Crane, this allows us to obtain $K$ corresponding text prompts `\texttt{Photo of a [sub-class] bird}', denoted by $\{\mathcal{\hat{P}}_1,\mathcal{\hat{P}}_2,...,\mathcal{\hat{P}}_{K} \}$.
%
%

\textbf{Retrieval-based Prompt.}
The prompt $\mathcal{\hat{P}}$ still is a handcrafted sentence template, we expect to develop it into a real language prompt in the human community.
One feasible solution for that is through prompt retrieval~\cite{beaumont-2022-clip-retrieval,radford2021learning}.
As shown in Fig.~\ref{pipeline}, given a prompt $\mathcal{\hat{P}}$, \ie{}, `\texttt{Photo of a [sub-class] car in the street}', Clipretrieval~\cite{beaumont-2022-clip-retrieval} pre-trained on Laion5B~\cite{schuhmann2022laion} is used to retrieve top $N$ real images and captions, where the captions as the final prompt sets.
Using this approach, we can collect a total of $K\times N$ text prompts, denoted by $\sum_{i=1}^{K\times N} \mathcal{\hat{P}}_i$, for our synthetic data. During inference, we randomly sample a prompt from this set to generate each image.

\begin{figure}[!t]
	\includegraphics[width=0.99\linewidth]{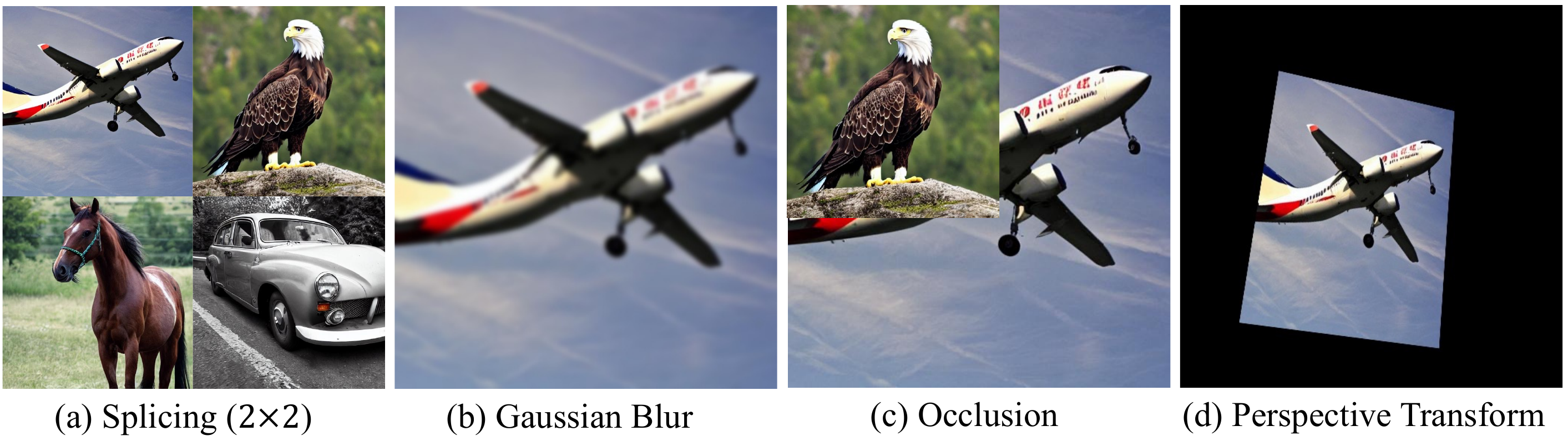}
	\vspace{-0.2cm}
	\caption{\textbf{Data Augmentation.} Four data augmentations are used to reduce the domain gap.}
	\vspace{-0.1cm}
\label{VIS12_data}
\end{figure}

\begin{table*}[t]
    \centering
    \small 
    \setlength{\tabcolsep}{1mm}
    \input{tables/VOC_semantic.tex}
    \vspace{-0.2cm}
    \caption{\textbf{Result of Semantic Segmentation on the VOC 2012 \texttt{val}.} \textit{mIoU} is for $20$ classes. `S' and `R' refer to `Synthetic' and `Real'.}
    \label{VOC_semantic}
\end{table*}


    

    

\begin{table}[t]
    \centering
    \small 
    \setlength{\tabcolsep}{1mm}
    \input{tables/cityscapes_human_car.tex}
    \vspace{-0.2cm}
    \caption{\textbf{The mIoU (\%) of Semantic Segmentation on Cityscapes \texttt{val}.} `Human' includes two sub-classes \texttt{person} and \texttt{rider}. `Vehicle' includes four sub-classes, \ie{}, \texttt{car}, \texttt{bus}, \texttt{truck} and \texttt{train}. Mask2former~\cite{cheng2022masked} with ResNet50 is used.}
    %
    
    
    \label{cityscapes_human}
\end{table}

\subsection{Data Augmentation}
\label{Data}
%
To further reduce the domain gap between the generated images and the real-world images in terms of size, blur, and occlusion,
data augmentations~$\Phi(\cdot )$~(\eg, Splicing~\cite{bochkovskiy2020yolov4}), as the effective strategies are used, as shown in Fig.~\ref{VIS12_data}.
\textbf{Splicing.} Synthetic image usually present normal size for the foreground~(object), \ie{}, objects typically occupy the majority of image.
However, real-world images often contain objects of varying resolutions, including small objects in datasets such as Cityscapes~\cite{(cityscape)cordts2016cityscapes}.
To address this issue, we use Splicing augmentation.
Fig.~\ref{VIS12_data} (a) presents one example for the image splicing~($2\times2$).
In the experiment, six scales of image splicing are used, \ie{}, $1\times2$, $2\times1$, $2\times2$, $3\times3$, $5\times5$, and $8\times8$, and the images are sampled from train set randomly. 
\textbf{Gaussian Blur.} Synthetic images typically exhibit a uniform level of blur, whereas real images exhibit varying degrees of blur due to motion, focus, and artifact issues.
Gaussian Blur~\cite{lopes2019improving} is used to increase the diversity of blur, where the length of Gaussian Kernel is randomly sampled from a range of $6$ to $22$.
\textbf{Occlusion}. Similar to CutMix~\cite{yun2019cutmix},  to make the model focus on discriminative parts of objects, patches of another image are cut and pasted among training images where the corresponding labels are also mixed
proportionally to the area of the patches.
\textbf{Perspective Transform.} Similar to the above augmentations, perspective transform is used to improve the diversity of the generated images by simulating different viewpoints.

\section{Experiments}
\subsection{Experimental Setups}
\textbf{Datasets and Task.} \textit{Datasets.} Following the previous works~\cite{cheng2022masked,li2023guiding} for semantic segmentation, Pascal-VOC 2012~\cite{(voc)everingham2010pascal},  ADE20k~\cite{(ade)zhou2017scene} and Cityscapes~\cite{(cityscape)cordts2016cityscapes} are used to evaluate \Ours. 
\textit{Tasks.} Three tasks are adopted in our experiment, \ie{}, semantic segmentation, open-vocabulary segmentation, and domain generalization.

\textbf{Implementation Details}
The pre-trained Stable Diffusion~\cite{rombach2022high}, the text encoder of CLIP~\cite{radford2021learning}, AffinityNet~\cite{ahn2018learning} are adopted as the base components.
We do not finetune the Stable Diffusion and only train AffinityNet for each category.
The corresponding parameter optimization and setting~(\eg{}, initialization, data augmentation, batch size, learning rate) all are similar to that of the original paper.
\textit{Synthetic data for training.} 
For each category on Pascal-VOC 2012~\cite{(voc)everingham2010pascal}, we generate $10k$ images and set $\alpha$ of noise learning to $0.7$ to filter $7k$ images.
As a result, we collect $60k$ synthetic data for $20$ classes as the final training set, and the spatial resolution is $512\times 512$.
For Cityscapes~\cite{cordts2016cityscapes}, we only evaluate $2$ important classes, \ie{}, `Human' and `Vehicle', including six sub-classes, \texttt{person}, \texttt{rider}, \texttt{car}, \texttt{bus}, \texttt{truck}, \texttt{train}, and generate $30k$ images for each sub-category, where $10k$ images are selected as the final training data by noise learning.
Considering the relationship between \texttt{rider} and \texttt{motorbike}/\texttt{bicycle}, we set the two classes to be ignored, while evaluating the `Human' class on Table~\ref{cityscapes_human} and Table~\ref{dg}.
In our experiment, only a single object for an image is considered.
Multi-categories generation~\cite{li2023guiding} usually causes the unstable quality of the images, limited by the generation ability of Stable Diffusion. 
Mask2Former~\cite{cheng2022masked} is used as the baseline to evaluate the dataset.
%
8 Tesla V100 GPUs are used for all experiments.

\textbf{Evaluation Metrics.} \textit{Mean intersection-over-union (mIoU)}~\cite{(voc)everingham2010pascal,cheng2022masked}, as the common metric of semantic segmentation, is used to evaluate the performance.
For open-vocabulary segmentation, following the prior~\cite{ding2022decoupling,cheng2021sign}, the mIoU averaged on seen classes, unseen classes, and their \textit{harmonic mean} are used.

\textbf{Mask Smoothness.} The mask $B_{\hat{\gamma}}$ generated by the Dense CRF often contains jagged edges and numerous small regions that do not correspond to distinct objects in the image.
To address these issues, we trained a segmentation model $\bm{\theta}$~(\ie{} Mask2Former), using the mask $B_{\hat{\gamma}}$ generated by the Dense CRF as input. 
We then used this model to predict the pseudo labels for the training set of synthetic data, resulting in a final semantic mask annotation

\textbf{Cross Validation for Noise Learning.}
In the experiment, we performed the three-fold cross-validation for each class.
The five-fold cross-validation (CV) is a process in which all data is randomly split into $k$ folds, in our case $k$ $=$ $3$, and then the model is trained on the $k - 1$ folds, while one fold is left to test the quality.

\begin{table}[t]
    \centering
    \small 
    \setlength{\tabcolsep}{1mm}
    \input{tables/zs3.tex}
    \vspace{-0.2cm}
    \caption{\textbf{Performance for Zero-Shot Semantic Segmentation Task on PASCAL VOC.} `Seen', `Unseen', and `Harmonic' denote mIoU of seen, unseen categories, and their harmonic mean. Priors are trained with real data and masks.}
    \label{zs3}
\end{table}

\begin{table*}[t]
  \input{tables/ablation_study.tex}
  \vspace{-0.2cm}
  \caption{\textbf{\diffmask ablations.} We perform ablations on VOC 2012 \texttt{val}. $\gamma$ and `AT' denotes the `Threshold' and `Adaptive Threshold', respectively.  $\alpha$ refers to the proportion of data pruning. $\Phi_1$, $\Phi_2$, $\Phi_3$ and $\Phi_4$ refer to `Splicing', `Gaussian Blur', `Occlusion', and `Perspective Transform', respectively. `Retri.' and `Sub-C' denotes `retrieval-based' and `Sub-Class', respectively. Mask2former with Swin-B is adopted as the baseline.
  }
  \label{ablation_diffmask}
  \vspace{-1mm}
\end{table*}
\subsection{Protocol-I: Semantic Segmentation}
\textbf{VOC 2012.} Table~\ref{VOC_semantic} presents the results of semantic segmentation on the VOC 2012.
%
%
The existing segmentation methods trained on synthetic data~(\diffmask) can achieve a competitive performance, \ie{}, $70.6\%$ $v.s.$ $84.3\%$ for mIoU with Swin-B backbone.
A point worth emphasizing is that our synthetic data does not need any manual localization and mask annotation, while real data need humans to perform a pixel-wise mask annotation. 
For some categories, \ie{}, \texttt{bird}, \texttt{cat}, \texttt{cow}, \texttt{horse}, \texttt{sheep}, \diffmask presents a powerful performance, which is quite close to that of training on real~(within $5\%$ gap).
Besides, finetune on few real data, the results can be improved further, and exceed that of training on full real data, \eg{}, $84.9\%$ mIoU finetune on $5.0$k real data $v.s$ $83.4\%$ mIoU training on full real data~($10.6$k).

\textbf{Cityscapes.} Table~\ref{cityscapes_human} presents the results on Cityscapes.
Urban street scenes of Cityscapes are more challenging, including a mass of small objects and complex backgrounds.
We only evaluate two classes, \ie{}, \texttt{Vehicle} and \texttt{Human}, which are the two most important categories in the driving scene.
Compared with training on real images, \diffmask presents a competitive result, \ie{}, $79.6\%$ $vs.$\  $90.8\%$ mIoU.

\textbf{ADE20K}
ADE20K, as one more challenging dataset, is also used to evaluate the \Ours.
Table~\ref{ADE20k_semantic} presents the results of three categories~(\texttt{bus}, \texttt{car}, \texttt{person}) on ADE20K.
With fewer synthetic images~($6$k), we achieve a competitive performance than that of a mass of real images~($20.2$k).
Compared with the other two categories, Class \texttt{car} achieves the best performance, with $73.4\%$ mIoU. 


\subsection{Protocol-II: Open-vocabulary Segmentation}
As shown in Fig.~\ref{fig:teaser}, it is natural and seamless to extend the text-driven synthetic data~(our \diffmask) to the open-vocabulary~(zero-shot) task.
%
%
As shown in Table~\ref{zs3}, compared with priors training on real images with manually annotated mask, \diffmask can achieve a SOTA result on \texttt{Unseen} classes.
It is worth mentioning that \diffmask is pure synthetic/fake data and supervised by text, while priors all must need the real image and corresponding manual mask annotation.
Li \textit{et al.}, as one contemporaneous work, use the segmentation model pre-trained on COCO~\cite{lin2014microsoft} to predict the pseudo label of the synthetic image, which is high-cost.

\begin{table}[t]
    \centering
    \small 
    \setlength{\tabcolsep}{1mm}
    \input{tables/ADE_semantic.tex}
    \vspace{-0.1cm}
    \caption{\textbf{The mIoU (\%) of Semantic Segmentation on the ADE20K \texttt{val}.}}
    \label{ADE20k_semantic}
\end{table}

\begin{table}[t]
    \centering
    \small 
    \setlength{\tabcolsep}{1mm}
    \input{tables/domain_generalization.tex}
    \vspace{-0.2cm}
    \caption{\textbf{Performance for Domain Generalization between different datasets.} Mask2former~\cite{cheng2022masked} with ResNet50 is used as the baseline. \texttt{Person} and \texttt{Rider} classes of Cityscapes~\cite{cordts2016cityscapes} are consider as the same class, \ie{}, \texttt{Person} in the experiment. }
    \label{dg}
\end{table}


\subsection{Protocol-III: Domain Generalization}
%
Table~\ref{dg} presents the results for cross-dataset validation, which can evaluate the generalization of data.
Compared with real data, \diffmask show powerful effectiveness on domain generalization, \eg{}, $69.5\%$ with \diffmask $v.s$ $68.0$ with ADE20K~\cite{(ade)zhou2017scene} on VOC 2012 \texttt{val}.
The domain gap~\cite{toldo2020unsupervised} between real datasets sometimes is bigger than that among synthetic and real data.
For \texttt{Motorbike} class, model training with Cityscapes only achieves $28.9\%$ mIoU, but that of \diffmask is $63.2\%$ mIoU.
We argue that the main reason is domain shift in foreground and background domains, \ie{}, Cityscapes contains images of city roads, with the majority of \texttt{Motorbike} objects being small in size.
But VOC 2012 is an open-set scenario, where \texttt{Motorbike} objects vary greatly in size and include close-up shots.

\subsection{Ablation Study}
\textbf{Compared with Attention Map.}
Table~\ref{ablation_DQ1} presents the comparison with the attention map and the impact of binarization threshold $\gamma$.
It is 
clear 
that the optimal threshold for different categories is different, even various for different images of the same category.
%
Sometimes it is sensitive for some categories, such as \texttt{Dog}.
The mIoU of $0.4$ $\gamma$ is better than that of $0.6$ $\gamma$ around $40\%$ mIoU, which can not be neglectful.
By contrast, our adaptive threshold is robust. 
Fig.~\ref{iou} also shows it is close to the optimal threshold.


\textbf{Prompt Engineering.} Table~\ref{tab:ablation:maskformer:b} provides the related ablation study for prompt strategies.
Retrieval-based and sub-classes prompt all can bring an obvious gain.
For \texttt{dog}, $10$ sub-classes prompt brings a $7.7\%$ mIoU improvement, which is quite significant.  
It is reasonable, the fine-grained prompts can directly enhance the diversity of generative images, as shown in Fig.~\ref{VIS12}.

\textbf{Noise Learning.} Table~\ref{ablation_NL} presents the impact of prune threshold $\alpha$.
$10k$ synthetic images for each class are used in this experiment.
%
%
The gain is 
considerable 
while $\alpha$ changes from $0.3$ to $0.5$.
In other experiments, we set the $\alpha$ to $0.7$ for each category.

\textbf{Data Augmentation.} The ablation study for the four augmentations is shown in Table~\ref{ablation_param}.
Compared with the other three augmentations, the gain of image splicing is the biggest.
One main reason is that the synthetic images are all $512\times512$ resolution and the size of the object usually is normal, image splicing can enhance the diversity of scale.

\textbf{What causes the performance gap between synthetic and real data.}
Domain gap and mask precision are the main reasons for the performance gap between synthetic and real data.
Table~\ref{gap} is set To further explore the problem.
Li \textit{et al.}~\cite{li2023guiding} shows that the pseudo mask of the synthetic image from Mask2former~\cite{cheng2022masked} pre-trained on VOC 2012 is quite accurate, and can as the ground truth.
Thus, we also use the pseudo label from the pre-trained Mask2former to train the model.
%
As shown in Table~\ref{gap}, mask precision cause $6.4\%$ mIoU gap, and the domain gap of images causes  $4.5\%$ mIoU gap.
Notably, for the \texttt{bird} class, the use of synthetic data with a pseudo label resulted in better results than the corresponding real images. 
This observation suggests that there may be no domain gap for the \texttt{bird} class in the VOC 2012 dataset.

\begin{table}[t]
    \centering
    \small 
    \setlength{\tabcolsep}{1mm}
    \input{tables/Ablation_Backbone}

    \vspace{-0.2cm}
    \caption{\textbf{Impact of Backbone on VOC 2012 \texttt{val}.} Mask2former~\cite{cheng2022masked} is used as the baseline. }
    \label{backbone}
\end{table}

\begin{table}[t]
    \centering
    \small 
    \setlength{\tabcolsep}{1mm}
    \input{tables/Ablation_GaP}

    \vspace{-0.2cm}
    \caption{\textbf{Impact of Mask Precision and Domain Gap on VOC 2012 \texttt{val}.} Mask2former~\cite{cheng2022masked} with Swin-B is used as the baseline. `Pseudo' denotes pseudo mask annotation from Mask2former~\cite{cheng2022masked} pre-trained on VOC 2012.}
    \label{gap}
\end{table}

\begin{figure}[!t]
	\includegraphics[width=0.99\linewidth]{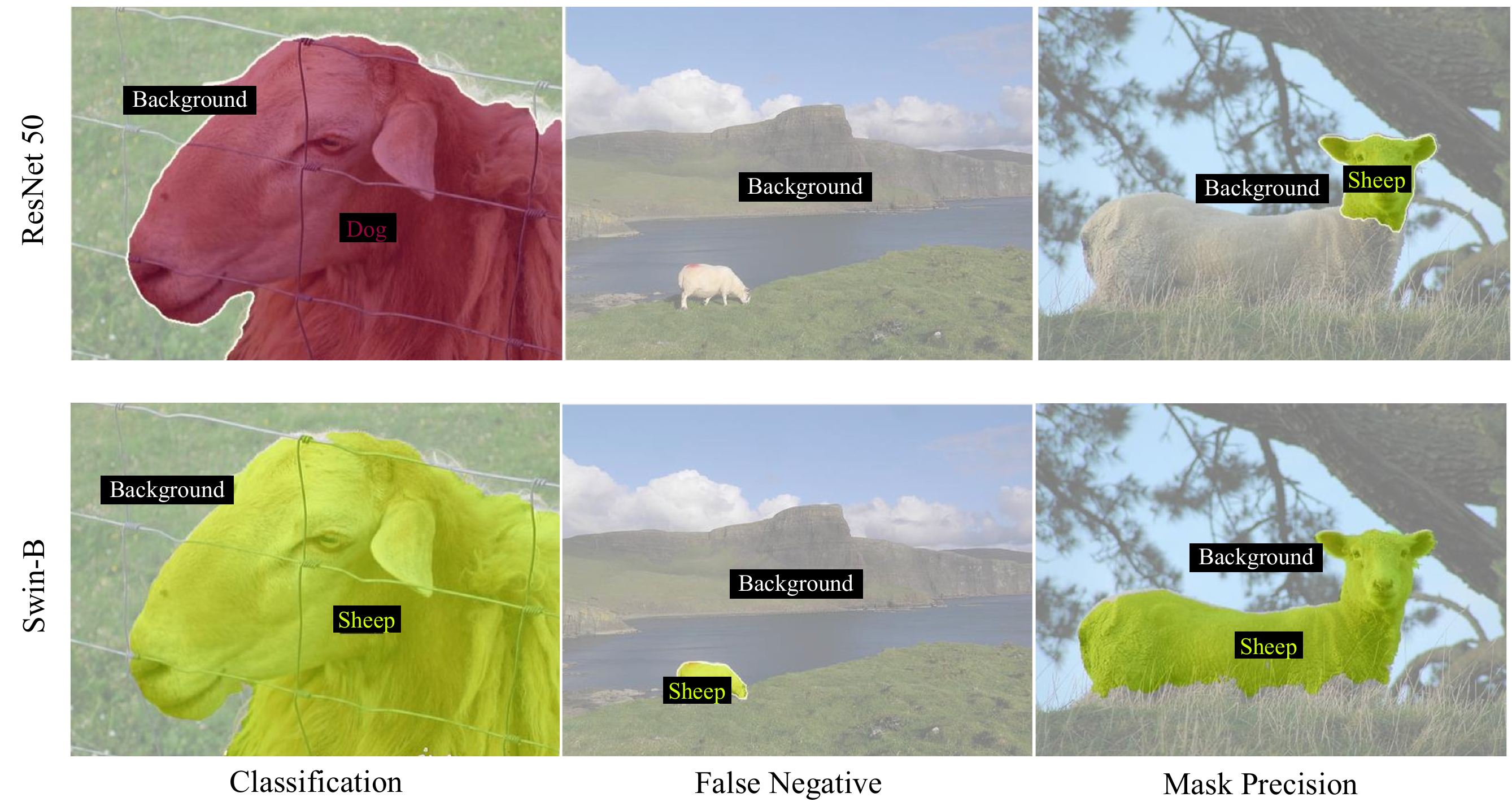}
	\vspace{-0.2cm}
	\caption{\textbf{Impact of Backbone.} Stronger backbone is robust for classification, False Negative, and mask precision.}
	\vspace{-0.1cm}
\label{backbone_vis}
\end{figure}

\textbf{Backbone}
Table~\ref{backbone} presents the ablation study for the backbone.
For some classes, \eg{} \texttt{sheep}, the stronger backbone can bring obvious gains, \ie{} Swin-B achieves $27.5\%$ mIoU improvement than that of ResNet 50.
And the mIoU of all classes with Swin-B achieves $19.2\%$ mIoU improvements.
It is an interesting and novel insight that a stronger backbone can reduce the domain gap between synthetic and real data. 
To give a further analysis for that, we present some results comparison of visualizations, as shown in Fig.~\ref{backbone_vis}.
Swin-B brings an obvious improvement in classification, False Negatives, and mask precision.
%

\section{Conclusion}
A new insight is presented in this paper, demonstrating that the accurate semantic mask of generative images can be automatically obtained through the use of a text-driven diffusion model.
%
To achieve this goal, we present \Ours, an automatic procedure to generate image and pixel-level semantic annotation.
The existing segmentation methods training on synthetic data of \Ours can achieve a competitive performance over the counterpart of real data.
Besides, \Ours shows the powerful performance for open-vocabulary segmentation, which can achieve a promising result on \texttt{Unseen} category.
We hope \diffmask can bring new insights and inspiration for bridging generative data and real-world data in the community.

\section*{Acknowledgements} 

W. Wu, C. Shen's participation was 
 supported by the National Key R\&D Program of China (No.\  2022ZD0118700). 
W. Wu, H. Zhou's participation was  supported by the National Key Research and Development Program of China (No.\ 2022YFC3602601), and the Key Research and Development Program of Zhejiang Province of China (No.\ 2021C02037).
M. Shou's participation was  supported by the National Research Foundation, Singapore under its NRFF Award NRF-NRFF13-2021-0008, and his Start-Up Grant from National University of Singapore.
Thank you to Runlong Liao for pointing out some citation errors.

{\small
\bibliographystyle{ieee_fullname}
\bibliography{main}
}
\end{document}

%% file: figures/fig1.tex
\twocolumn[{
\renewcommand\twocolumn[1][]{#1}%
\maketitle
\begin{center}
    \centering
  	\captionsetup{type=figure}
	\includegraphics[width=0.98\linewidth]{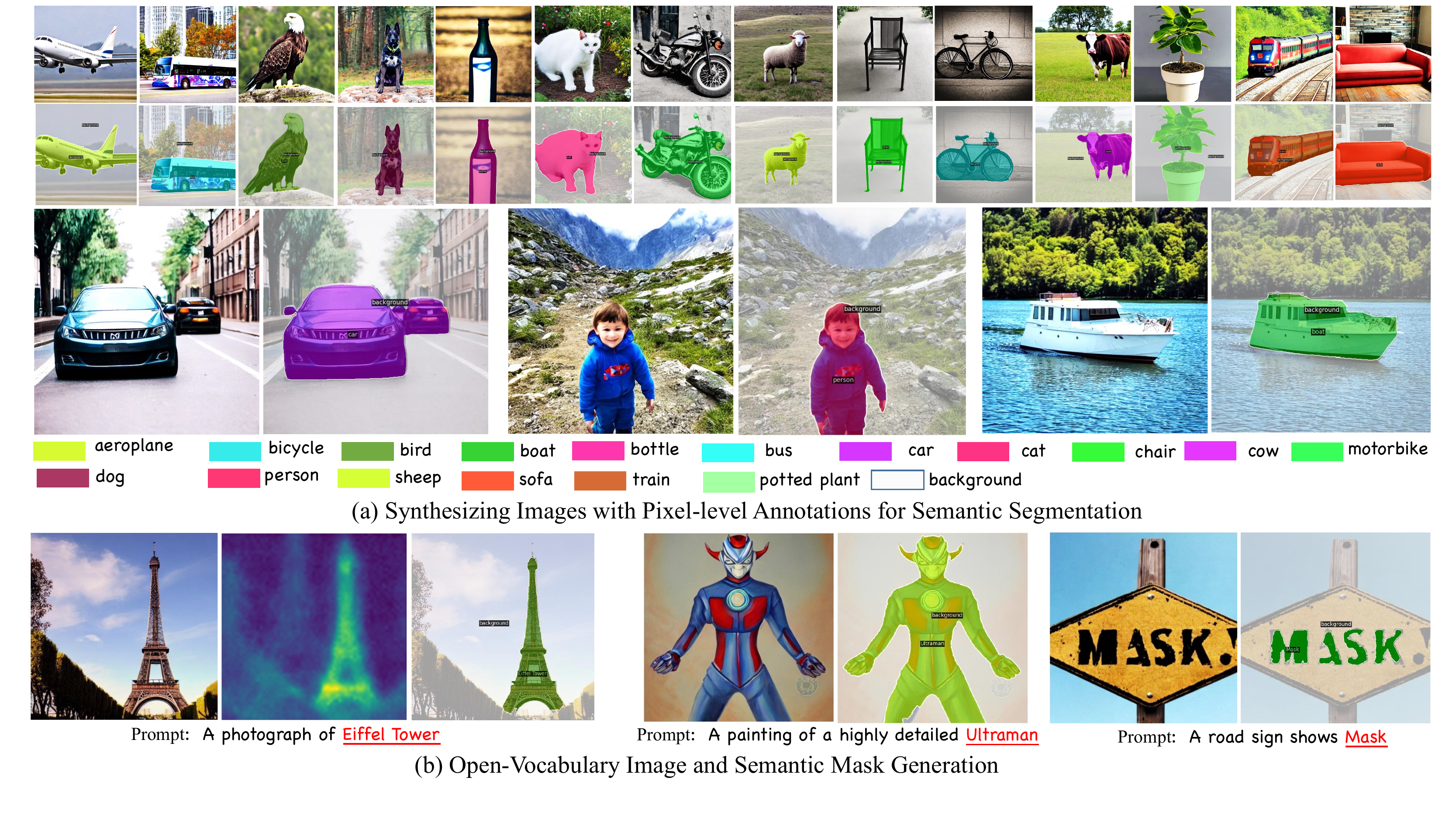}\\[-3mm]
	\captionof{figure}{\textbf{\diffmask synthesizes photo-realistic images 
    and high-quality mask  
    annotations 
    by exploiting the 
    attention maps of the diffusion model.} 
    Without human effort 
    for 
    localization 
    \diffmask 
    is capable of 
    producing  high-quality semantic masks.}
	\label{fig:teaser}

\end{center}
}]

%% file: tables/VOC_semantic.tex
\tablestyle{4pt}{1.2}\scriptsize\begin{tabular}{l|c |c | ccccccccccccc | c}
  & & & \multicolumn{12}{c}{Semantic Segmentation~(IoU) for Selected Classes/\%} & \\
Train Set & Number & Backbone & aeroplane & bird & boat & bus & car & cat & chair & cow & dog & horse  & person & sheep & sofa  & mIoU \\
  \shline
  \multicolumn{17}{l}{\textit{Train with Pure \textbf{Real} Data}}

  \\
  \multirow{2}{*}{VOC}
  & R: 10.6k (all) & R50 &  87.5 & 94.4 & 70.6 &  95.5 & 87.7 & 92.2 & 44.0 & 85.4 &  89.1 & 82.1 & 89.2 & 80.6 & 53.6 & 77.3  \\
  & R: 10.6k (all) & Swin-B & 97.0 & 93.7 & 71.5 &  91.7 & 89.6 & 96.5 & 57.5 & 95.9 &  96.8 & 94.4 & 92.5 & 95.1 & 65.6  & 84.3  \\
  & R: 5.0k  & Swin-B &  95.5 &  87.7 & 77.1  &  96.1   & 91.2 &  95.2 & 47.3 & 90.3  & 92.8  & 94.6  & 90.9 & 93.7  &  61.4 &   83.4  \\
   \hline
   \multicolumn{17}{l}{\textit{Train with Pure \textbf{Synthetic} Data}}\\
  \multirow{2}{*}{\textbf{\diffmask}}
  & S: 60.0k & R50\phantom{$^{\text{\textdagger}}$} & 80.7 & 86.7 & 56.9 & 81.2 & 74.2 &  79.3 & 14.7 & 63.4 &  65.1 & 64.6 & 71.0 & 64.7 & 27.8  & 57.4  \\
  & S: 60.0k & Swin-B\phantom{$^{\text{\textdagger}}$} & 90.8 & 92.9 & 67.4 & 88.3 & 82.9 &  92.5 & 27.2 & 92.2 & 86.0 & 89.0 &  76.5 & 92.2 & 49.8  & 70.6  \\
   \hline
   \multicolumn{17}{l}{\textit{Finetune on  \textbf{Real} Data}}\\
  \multirow{2}{*}{VOC, \textbf{\diffmask}}
  & S: 60.0k + R: 5.0k & R50\phantom{$^{\text{\textdagger}}$} & 85.4 & 92.8 & 74.1 & 92.9 & 83.7 & 91.7& 38.4 & 86.5 & 86.2 & 82.5 & 87.5 & 81.2 & 39.8  & 77.6 \\
  & S: 60.0k + R: 5.0k & Swin-B\phantom{$^{\text{\textdagger}}$} & 95.6 & 94.4 & 72.3 & 96.9 & 92.9 & 96.6 & 51.5 & 96.7 &  95.5 & 96.1 & 91.5 & 96.4 & 70.2 & 84.9 \\
  
  \end{tabular}

%% file: tables/cityscapes_human_car.tex
\tablestyle{4pt}{1.2}\scriptsize\begin{tabular}{l|c |c |cc|c}
    & & & \multicolumn{2}{c|}{Category/\%} & \\
Train Set & Number & Backbone & Human & Vehicle  & mIoU \\
  \shline
  \multicolumn{6}{l}{\textit{Train with Pure \textbf{Real} Data}}\\
  \multirow{2}{*}{Cityscapes}
  & 3.0k (all) & R50 &  83.4 & 94.5  & 89.0   \\
  & 3.0k (all) & Swin-B & 85.5 & 96.0  &  90.8  \\
  & 1.5k & Swin-B & 84.6 & 95.3  &  90.0  \\
   \hline
   \multicolumn{6}{l}{\textit{Train with Pure \textbf{Synthetic} Data}}\\
  \multirow{2}{*}{\textbf{\diffmask}}
  & 100.0k & R50\phantom{$^{\text{\textdagger}}$} & 70.7 & 85.3   & 78.0\\
  & 100.0k & Swin-B\phantom{$^{\text{\textdagger}}$} & 72.1 &  87.0 & 79.6 \\
   \hline
   \multicolumn{6}{l}{\textit{Finetune with \textbf{Real} Data}}\\
  \multirow{2}{*}{Cityscapes, \textbf{\diffmask}}
  & 100.0k + 1.5k & R50\phantom{$^{\text{\textdagger}}$} & 84.6 & 95.5 &  90.1 \\
  
  & 100.0k + 1.5k & Swin-B\phantom{$^{\text{\textdagger}}$} & 86.4 &  96.4 & 91.4  \\
  
  \end{tabular}


%% file: tables/zs3.tex
\tablestyle{4pt}{1.2}\scriptsize\begin{tabular}{l|cc|ccc}
    &  \multicolumn{2}{c|}{Train Set/\%} &  \multicolumn{3}{c}{mIoU/\%}  \\
  Methods & Type & Categories & Seen & Unseen & Harmonic \\
  \shline
  \multicolumn{6}{l}{\textit{Manual \textbf{\underline{Mask}} Supervision}}\\
  ZS3~\cite{bucher2019zero}            &real& 15  & 78.0        & 21.2   & 33.3     \\
  CaGNet~\cite{gu2020context}         &real& 15  & 78.6        & 30.3   & 43.7     \\
  Joint~\cite{baek2021exploiting}       & real &  15  & 77.7        & 32.5   & 45.9     \\
  STRICT~\cite{pastore2021closer}     &real&   15   & 82.7        & 35.6   & 49.8     \\
  SIGN~\cite{cheng2021sign}           &real& 15  &  \underline{83.5}       & 41.3   & 55.3     \\
  ZegFormer \cite{ding2022decoupling} &real & 15 &   \textbf{86.4}        & \underline{63.6}   & \textbf{73.3} \\

  \shline
  \multicolumn{6}{l}{\textit{Pseudo \textbf{\underline{Mask}} Supervision from Model pre-trained on COCO~\cite{lin2014microsoft}}}\\
  
  Li \textit{et al.}~\cite{li2023guiding}~(ResNet101) & synthetic &15+5& 62.8          &    50.0    &  55.7 \\
  \shline
  \multicolumn{6}{l}{\textit{\textbf{\underline{Text(Prompt)}}} Supervision} \\
  \diffmask~(ResNet50) & synthetic & 15+5 & 60.8 & 50.4 & 55.1   \\
  \diffmask~(ResNet101) & synthetic & 15+5 & 62.1 & 50.5 & 55.7   \\
  \diffmask~(Swin-B) & synthetic & 15+5 & 71.4 & \textbf{65.0} & \underline{68.1}  
  
  \end{tabular}


%% file: tables/ablation_study.tex
  
  \begin{subtable}{0.24\linewidth}
  \centering
  \tablestyle{2.3pt}{1.2}
  \footnotesize
  \begin{tabular}{ r |c|cc|c}
   Annotation & $\gamma$& Bird & Dog & \textit{mIoU} \\
  \shline
   Affinity map & - & 84.4  & 78.8  & 81.6 \\
   Attention & 0.4 &  88.1 & 82.4 & 85.3  \\
   Attention & 0.5 &  90.3 & 67.4 & 78.9 \\
   Attention & 0.6 &  50.5 & 38.3 & 44.4 \\
   \diffmask & AT & 92.9  & 86.0 &  89.5
  \end{tabular}
  \caption{\textbf{\diffmask $v.s.$ Attention Map}.
  }
  \label{ablation_DQ1}
  \end{subtable}\hspace{0.5mm}
  \begin{subtable}{0.23\linewidth}
  \centering
  \tablestyle{2.3pt}{1.2}
  \footnotesize
  \begin{tabular}{cc|cc|c}
  Retri. & Sub-C &  Bird & Dog & \textit{mIoU} \\
  \shline
  &  & 78.2 & 75.6 & 76.9\\
  \checkmark &  & 79.2 & 76.2 &77.7\\
   \checkmark & 10  & 91.3 & 83.9 &87.6\\
   \checkmark& 50  & 92.5 & 85.4 &89.0\\
   \checkmark& 100  & 92.9  & 86.0 &  89.5
  \end{tabular}
  \caption{\textbf{Prompt Engineering.} 
  }
  \label{tab:ablation:maskformer:b}
  \end{subtable}\hspace{0.5mm}
  \begin{subtable}{0.22\linewidth}
  \centering
  \tablestyle{3pt}{1.2}
  \footnotesize
  \begin{tabular}{l|cc|c}
  $\alpha$ & Bird & Dog & \textit{mIoU} \\
  \shline
  0.3  & 87.2 & 79.2 & 83.2\\
  0.4  & 89.5 & 79.9 & 84.7\\
  0.5 & 91.9 & 84.4 & 88.2\\
  0.6 & 92.6 & 85.2 & 89.1\\
  0.7 & 92.9  & 86.0 &  89.5
  
  \end{tabular}
  \caption{\textbf{Noise Learning.}
  }
  \label{ablation_NL}
  \end{subtable}
  \hspace{0.5mm}
  \begin{subtable}{0.25\linewidth}
  \centering
  \tablestyle{3pt}{1.2}
  \footnotesize
  \begin{tabular}{l|cc|c}
  Method & Bird & Dog  &\textit{mIoU} \\
  \shline
  $-$   & 87.0 & 81.5 & 84.3 \\
  $\Phi_1$  & 90.2 &83.7 & 87.0\\
  $\Phi_1$, $\Phi_2$ & 90.9 & 84.8 &87.9 \\
  $\Phi_1$, $\Phi_2$, $\Phi_3$ & 91.2 & 85.1 & 88.2\\
  $\Phi_1$, $\Phi_2$, $\Phi_3$,  $\Phi_4$ & 92.9  & 86.0 &  89.5
  \end{tabular}
  \caption{\textbf{Data Augmentation.}
  }
  \label{ablation_param}
  \end{subtable}

%% file: tables/ADE_semantic.tex
\tablestyle{4pt}{1.2}\scriptsize\begin{tabular}{l|l |l | ccc | c}
  & & & \multicolumn{3}{c}{Category/\%} & \\
Train Set & Number & Backbone  & bus & car     & person   & mIoU \\
  \shline
  
  \multicolumn{7}{l}{\textit{Train with Pure \textbf{Real} Data}}\\
  \multirow{2}{*}{ADE20K} & R: 20.2k & R50  & 87.9  & 82.5 &  79.4 &  83.3 \\
  & R: 20.2k & Swin-B  &  93.6 & 86.1  & 84.0 &   87.9   \\
   \hline
   \multicolumn{7}{l}{\textit{Train with Pure \textbf{Synthetic} Data}}\\
  \multirow{2}{*}{\textbf{\diffmask}}
  & S: 6.0k & R50\phantom{$^{\text{\textdagger}}$}  & 43.4 & 67.3 &  60.2 &  57.0     \\
  & S: 6.0k & Swin-B\phantom{$^{\text{\textdagger}}$}  & 72.8 & 73.4 &  62.6 &  69.6     \\
  
  \end{tabular}

%% file: tables/domain_generalization.tex
\tablestyle{4pt}{1.2}\scriptsize\begin{tabular}{ r |c|ccc|c}
   & &  \multicolumn{4}{c}{mIoU/\%}  \\
  Train Set &  Test Set & Car & Person & Motorbike & mIoU\\
  \shline
  Cityscapes~\cite{cordts2016cityscapes} & VOC 2012~\cite{(voc)everingham2010pascal} \texttt{val}&  26.4 &  32.9  & 28.3    &     29.2    \\
  ADE20K~\cite{(ade)zhou2017scene}      & VOC 2012~\cite{(voc)everingham2010pascal} \texttt{val} &  73.2 & 66.6        & \textbf{64.1}       & 68.0\\
  \diffmask   & VOC 2012~\cite{(voc)everingham2010pascal} \texttt{val} &  \textbf{74.2} &    \textbf{71.0}     & 63.2      & \textbf{69.5} \\
  \shline
  VOC 2012~\cite{(voc)everingham2010pascal}     &  Cityscapes~\cite{cordts2016cityscapes} \texttt{val}&    \textbf{85.6}       &  53.2    &  11.9 & 50.2\\
  ADE20K~\cite{(ade)zhou2017scene}       & Cityscapes~\cite{cordts2016cityscapes} \texttt{val}&    83.3       &  63.4  & \textbf{33.7} & \textbf{60.1} \\
  \diffmask       & Cityscapes~\cite{cordts2016cityscapes} \texttt{val}& 84.0 &     \textbf{70.7}    &  23.6 & 59.4
  
  \end{tabular}


%% file: tables/Ablation_Backbone.tex
\tablestyle{4pt}{1.2}\scriptsize\begin{tabular}{l|ccccc|c}
   Backbone & Bird & Dog & Sheep & Horse & Person &\textit{mIoU} \\
  \shline
   RseNet 50 & 86.7   & 65.1 &  64.7 & 64.6 & 71.0 & 70.3\\
   RseNet 101 & 86.7 & 66.8  & 65.3 & 63.4& 70.2& 70.5\\
   Swin-B & 92.9  & 86.0 & 92.2 & 89.0 & 76.5 & 87.3 \\
   Swin-L & 92.8  & 86.4 & 92.3 & 88.3 & 77.3& 87.4

  \end{tabular}

%% file: tables/Ablation_GaP.tex
\tablestyle{4pt}{1.2}\scriptsize\begin{tabular}{l|cccc|c}
   Annotation & Bird & Dog & Person & Sofa & \textit{mIoU} \\
  \shline
   Real Image, Manual Label & 93.7   & 96.8 & 92.5& 65.6& 87.2\\
   Synthetic Image, Pseudo Label & 95.2   & 86.2 & 89.9 & 59.5& 82.7 \\
   Synthetic Image, \diffmask & 92.9  & 86.0 &   76.5 & 49.8 & 76.3

  \end{tabular}